\documentclass[journal]{IEEEtran}

\usepackage{subfigure}
\usepackage{multirow}
\usepackage{mathtools}
\usepackage{amssymb}
\usepackage{color}

\begin{document}
	
	
	
	\title{Self-directed Machine Learning}
	
	
	
	\newcommand{\eg}{\textit{e.g.}}

	\author{Wenwu Zhu,~\IEEEmembership{Fellow,~IEEE,}
		Xin~Wang,~\IEEEmembership{Member,~IEEE,}
		and~Pengtao~Xie
		\thanks{Wenwu Zhu and Xin Wang are with the Department
			of Computer Science and Technology, Tsinghua University, China,
			e-mail: \{wwzhu, xin\_wang\}@tsinghua.edu.cn}
		\thanks{Pengtao Xie is with the Department of Electrical and Computer Engineering, University of California, San Diego, USA, e-mail: p1xie@ucsd.edu}
		\thanks{Corresponding Author: Xin Wang}
	}

	
	\markboth{Journal of \LaTeX\ Class Files,~Vol.~14, No.~8, August~2015}%
	{Shell \MakeLowercase{\textit{et al.}}: Bare Demo of IEEEtran.cls for IEEE Journals}
	
	\maketitle
	\begin{abstract}
		
		Conventional machine learning (ML) relies heavily on manual design from machine learning experts to decide learning tasks, data, models, optimization algorithms, and evaluation metrics, which is labor-intensive, time-consuming, and cannot learn autonomously like humans. In education science, self-directed learning, where human learners select learning tasks and materials on their own without requiring hands-on guidance, has been shown to be more effective than passive teacher-guided learning. Inspired by the concept of self-directed human learning, we introduce the principal concept of Self-directed Machine Learning (SDML) and  propose a framework for SDML.
		Specifically, we design SDML as a self-directed learning process guided by self-awareness, including internal awareness and  external awareness. Our proposed SDML process benefits from self task selection, self data selection, self model selection, self optimization strategy selection and self evaluation metric selection through self-awareness without human guidance.   
		Meanwhile, the learning performance of the SDML process serves as feedback to further improve self-awareness. We propose a mathematical formulation for SDML based on multi-level optimization. 
		Furthermore, we present case studies together with potential applications of SDML, followed by  discussing future research directions. We expect that SDML could enable machines to conduct human-like self-directed learning and provide a new perspective towards artificial general intelligence.
	\end{abstract}
	
	\IEEEpeerreviewmaketitle
	\section{Introduction}
	
	
	Machine learning has achieved substantial 
	success in many areas such as natural language processing, computer vision, and robotics. 
	Towards the ultimate goal of artificial general intelligence (AGI), researchers have kept working on reducing manual design in  learning processes.  Unsupervised learning, semi-supervised learning and self-supervised learning methods have shown their strength~\cite{grira2004unsupervised, qi2020small, Lars2020asurvey, jaiswal2020survey, liu2020self} in reducing manual annotations, and active learning can help to reduce the cost for labeling data by interactively querying users or some other information sources to label ``important'' new data points~\cite{wang2011active, kumar2020active, ren2020survey}. Besides, automated machine learning (AutoML) carries out neural architecture search and  hyperparameter optimization~\cite{yao2018taking, he2019automl, zoller2019survey} for the sake of reducing manual efforts in  model design and selection. 
	Meta learning utilizes a meta learner to quickly adapt machine learning algorithms to new tasks with a small amount of new data without manually  transferring knowledge~\cite{finn2017model, vanschoren2018meta, hospedales2020meta}.  
	
	Despite all the above progress, 
	current machine learning paradigm is still heavily dependent on manual design and human guidance, 
	where human experts 
	decide  learning tasks, data, models, optimization algorithms  and evaluation metrics. For example, 
	to develop a rescue robotics system, human experts need to design how to  
	train analogous reasoning models on 
	manually-selected  datasets, how to perform automatic knowledge graph construction, what evaluation metrics  to use, etc. 
	Manual design is labor-intensive, time-consuming, and lacks autonomy.  
	To address this problem, we are interested in investigating  whether it is possible  to  let 
	machines select learning   tasks, data, models,  optimization algorithms,  evaluation metrics autonomously and control  learning processes  in a self-directed manner?
	
	Self-directed human learning (SDHL) has been studied in education science for many years since the 1970s~\cite{knowles1975self, garrison1997self, caffarella1993self}.  Particularly, in adult education~\cite{loeng2020self,brookfield1993self} and online education~\cite{song2007conceptual, latour2021self}, where learning initiatives of learners are strong required, 
	self-directed human learning shows great effectiveness in improving learning outcomes. 
	In self-directed human learning,  individuals take the initiative to   determine proper learning tasks, selecting appropriate learning materials, making suitable learning plans, developing  effective learning strategies, and  adopting effective  evaluation metrics. It has been shown that self-directed human learners who take the initiative in learning tend to learn more and better than those who  passively wait for guidance from teachers~\cite{knowles1975self}. 

	\begin{figure*}[h!t!]
		\centering
		\begin{tabular}{cc}
			\hspace{-0mm}\includegraphics[width=.45\textwidth, height = 5cm]{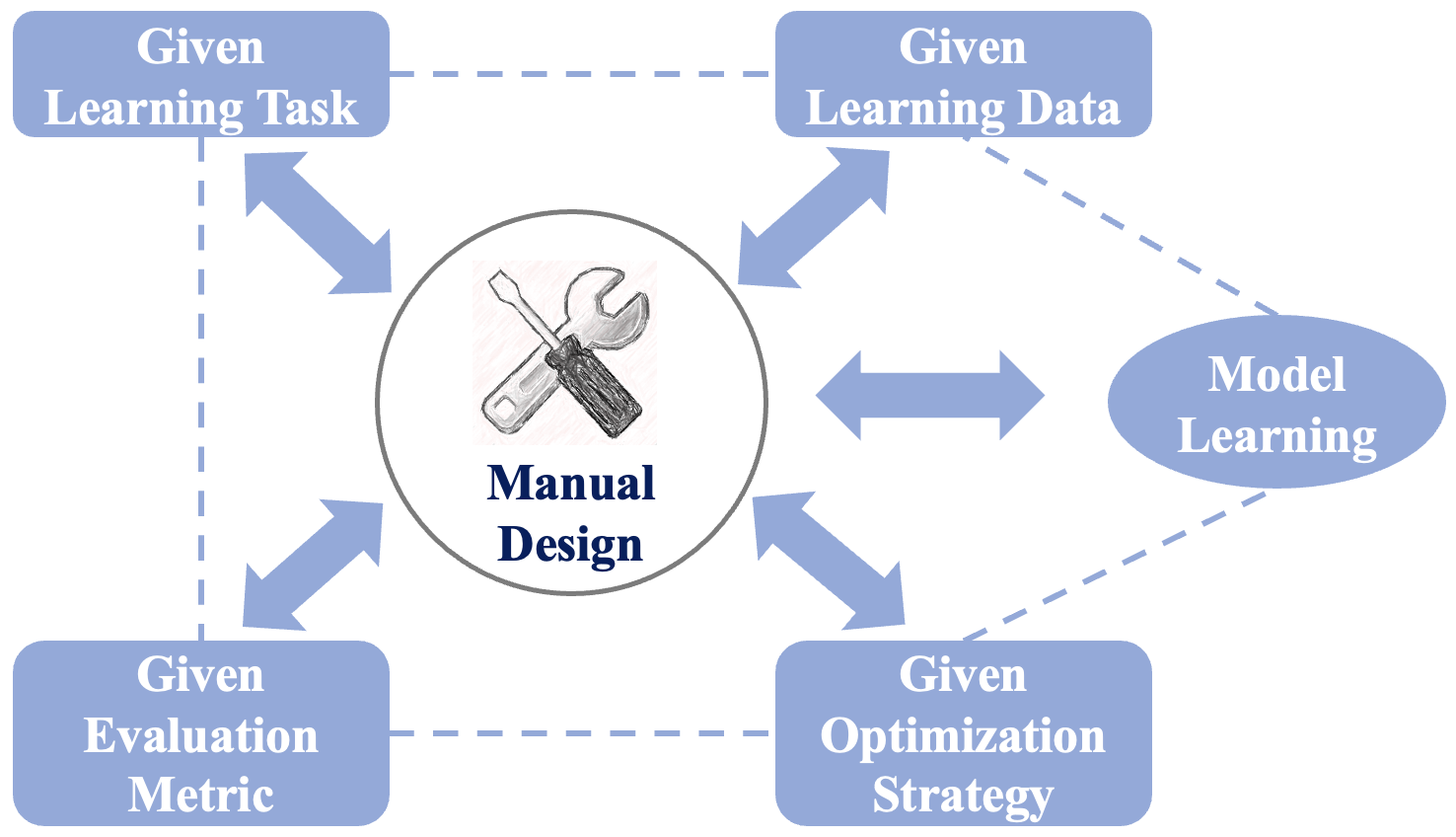} &
			\hspace{-0mm}\includegraphics[width=.45\textwidth, height = 5cm]{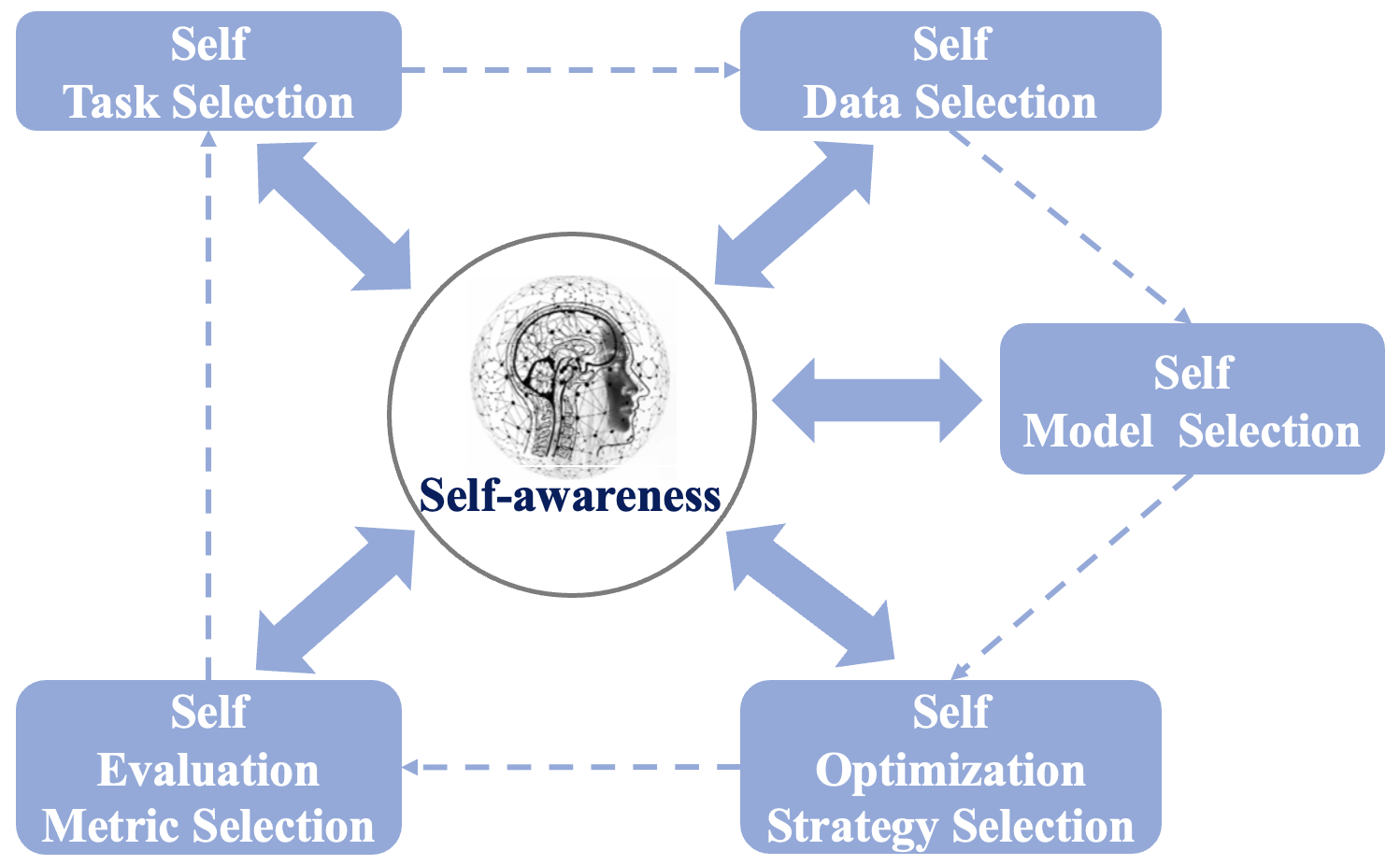} \\
			\hspace{-0mm}(a) Conventional Machine Learning & \hspace{-0mm}(b) Self-directed Machine Learning 
		\end{tabular}
		\caption{Comparison of conventional machine learning (a) and self-directed machine learning (b).}
		\label{fig:comp}
	\end{figure*}

	\begin{figure*}[!ht]
		\centering
		\includegraphics[width=.8\linewidth]{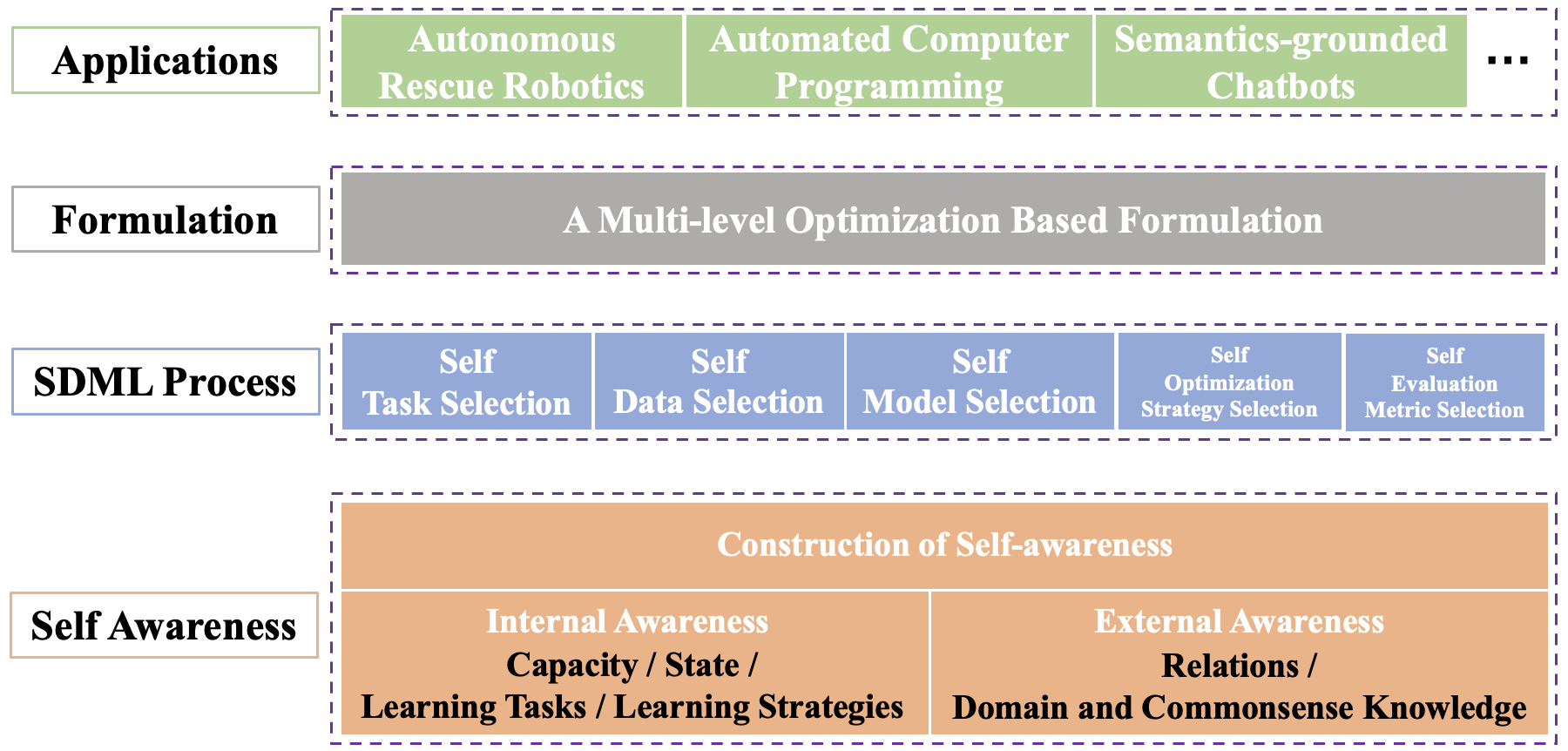}
		\caption{Illustration diagram of Self-directed Machine Learning (SDML).}
		\label{fig:archi}
	\end{figure*}
	
	Inspired by the concept of self-directed human learning, we propose \textbf{self-directed machine learning (SDML)}.
	The comparison between conventional machine learning and SDML is illustrated in Fig.~\ref{fig:comp}. While conventional machine learning mostly depends on  manual designs from humans, SDML leverages \textit{self-awareness} to   perform a self-directed learning process with high degree of autonomy, where the learning process in turn can further improve  \textit{self-awareness}. 
	Self-awareness includes internal awareness and external awareness. The internal awareness reveals the states of machines themselves, while the external awareness represents machines' perception abilities towards the physical world and how they are seen by the external world. With the guidance of self-awareness, SDML is able to conduct a  self-directed learning process through self task selection, self data selection, self model selection, self optimization algorithm  selection and self evaluation metric selection in a more self-directed way. We propose a mathematical formulation for SDML based on multi-level optimization.  Moreover,  we present case studies together with potential
	applications of SDML, followed by discussing future research
	directions. 
	
	
	The rest of the paper is organized as follows. Section II reviews self-directed human learning. In  Section III, we introduce the principal concept and propose the framework of self-directed machine learning. 
	In Section~\ref{sec:case} we  showcase the superiority of SDML over conventional machine learning by discussing several case studies and potential applications for SDML, such as autonomous driving/robotics and automated computer programming. We discuss future directions in Section~\ref{sec:concl}, including equipping SDML with robustness, explainability and reasoning capability.

	\section{Self-directed Human Learning}
	
	The concept of self-directed human learning (SDHL) has been studied in  educational science for decades~\cite{knowles1975self, garrison1997self, caffarella1993self}, especially in adult education~\cite{loeng2020self,brookfield1993self} and online education~\cite{song2007conceptual, latour2021self} as the two scenarios both require high initiative of learners. The original definition of self-directed learning~\cite{knowles1975self} is provided by Knowles in 1975:
	
	\textit{In its broadest meaning, self-directed learning describes a process by which individuals take the initiative, with or without the assistance of others, in diagnosing their learning needs, formulating learning goals, identifying human and material resources for learning, choosing and implementing appropriate learning strategies, and evaluating learning outcomes.} 
	It has been shown that  self-directed learners who take the initiative can learn more and better than those who  passively learn under guidance from teachers~\cite{knowles1975self}, and can know how they see themselves and how they are seen by others. 
	
	
	Self-directed learning relies on self-awareness.   
	In education science~\cite{eurich2017insight}, there are two types of self-awareness: internal awareness and external awareness. 
	By combining internal awareness and external awareness, learners can  identify what to  improve  and change the  way  of interacting with themselves, with others and with the physical world. As such, being able to master both internal and external awareness is regarded as a very crucial prerequisite for  conducting self-directed (human) learning.

	\section{Self-directed Machine Learning}
	\label{sec:SDML}
	We propose a new machine learning concept and framework called  self-directed machine learning (SDML), inspired by the concept of SDHL in education science. Conventional machine learning is mostly human-directed, which could handle well-defined specific tasks but is unable to automatically adapt to changing environments. 
	In contrast, SDML takes the initiative in  learning processes and pursues lifelong self-improvement, as shown in Fig.~\ref{fig:comp}b and Fig.~\ref{fig:archi}.
	
	In education science, self-directed learning views learners as responsible owners and managers of their own learning process~\cite{abdullah2001self}. 
	Internal awareness and external awareness are treated as two important components for improving and changing learners'  perceptions about themselves and others. Drawing inspirations from these facts, we design SDML as a self-directed learning process guided by self-awareness, including internal 
	awareness and external awareness.  During the learning process, SDML is  towards  being self-directed to select learning tasks, data, models, optimization algorithms, and evaluation metrics. In turn, learning outcomes 
	provide feedback on how to improve  self-awareness.
	
	\subsection{Self-awareness}
	
	\begin{figure}[!ht]
		\centering
		\includegraphics[width=0.9\columnwidth]{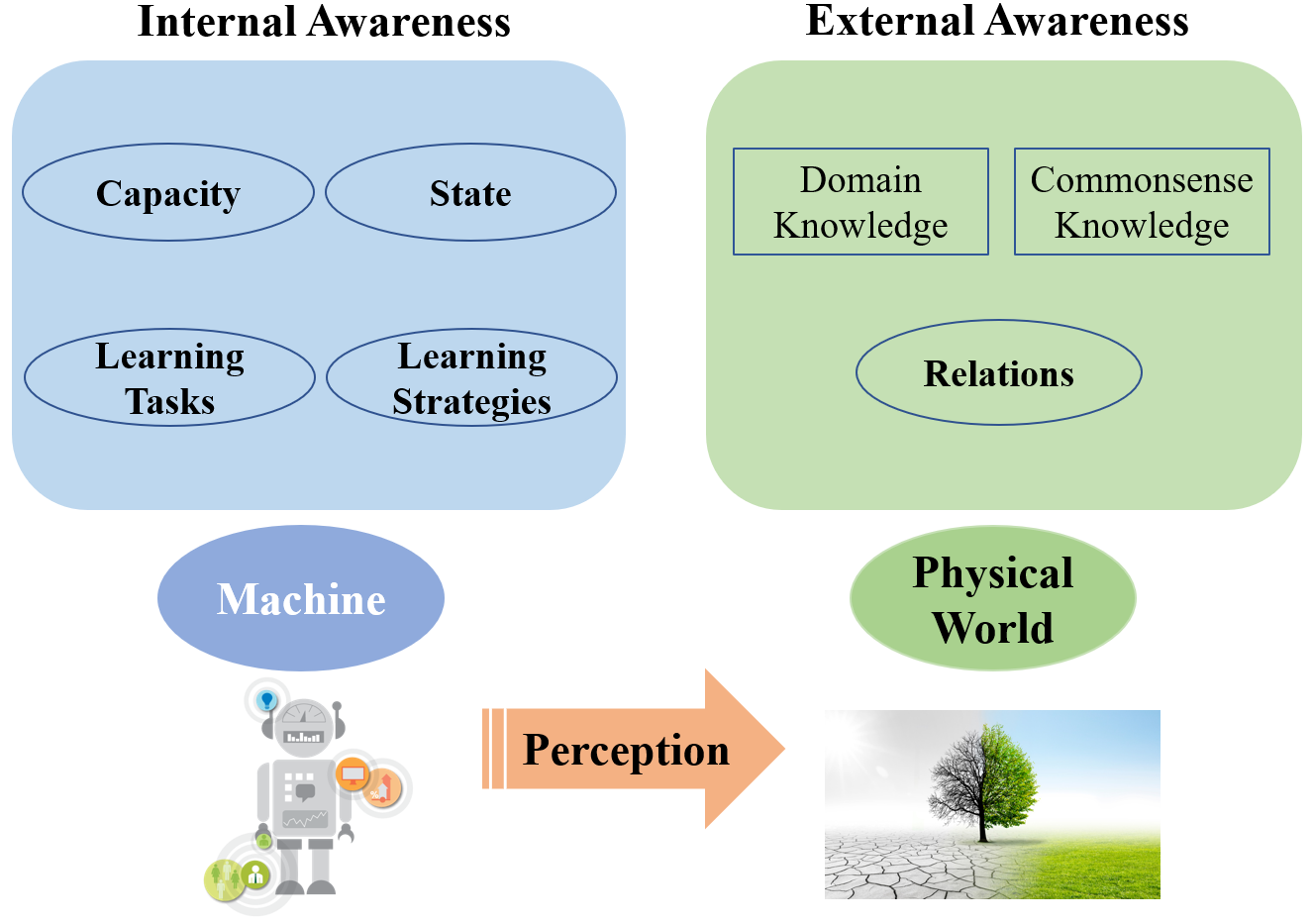}
		\caption{Self-awareness: i) Internal Awareness; ii) External Awareness}
		\label{fig:learner}
	\end{figure}
	
	
	Motivated by discoveries in education science~\cite{schunk2012motivation,eurich2017insight}, we propose a way of implementing self-awareness, including internal awareness and external awareness, as shown in Fig.~\ref{fig:learner}.
	
	
	\subsubsection{Internal Awareness}

	
	As SDML requires  
	lifelong  
	self-improvement of machines, it is essential to build machines'  internal awareness to better guide  the learning process. Inspired by  humans' cognitive conditions in self-directed learning, we represent internal awareness with the following key factors: 
	\begin{enumerate}
		\item \textit{Capacity}. 
		We represent the long-term internal awareness of a machine using capacity $\mathbb{C}\coloneqq\{C_1, C_2, ..., C_K\}$, where $\{C_{i}\}_{i=1}^K$ are different aspects of capacity such as computation speed and memory size. Machine capacity specifies the 
		maximum available resources of an SDML system.
		\item \textit{State}. 
		We represent  the short-term internal awareness of a machine using state $\mathbb{S}\coloneqq\{S_1, S_2, ..., S_K\}$, where $\{S_{i}\}_{i=1}^K$ are the current states of the corresponding capacity, $\{C_{i}\}_{i=1}^K$, such as current computation speed and currently available memory size. Machine state represents the current ability and available resources of an SDML system.
		
		\item \textit{Learning Tasks}. The COPES model of learning tasks includes Conditions, Operations, Products, Evaluations, and Standards~\cite{winne1989theories, winne1997experimenting}. Inspired by this, 
		we define a machine learning task as $\{T_C, T_O, T_E, T_S\}$, where $T_C$ is task conditions on inputs and system constraints, $T_O$ is task outputs (products), $T_E$ is evaluations and $T_S$ is standards. As machine learning tasks are related (e.g., VQA tasks are supported by visual and textual representation learning tasks), we organized the set of known tasks as graphs $G_T=\{\mathbb{T}, \mathbb{E}\}$, where $\mathbb{T}$ is the set of machine learning tasks and $\mathbb{E}$ is their relations. $G_T$ can be initialized with prior knowledge and updated during the learning process/from external resources.
		\item \textit{Learning Strategies}. People can develop various and personalized study strategies for different scenarios in their lifetime. We denote the set of machine learning strategies as $\mathbb{Q}$, which consists of strategies for model constructions, parameters optimizations, evaluation metrics and so on. It is worth noting that $\mathbb{Q}$ includes the models and parameters pre-trained or already trained in the machine learning task set $\mathbb{T}$.
	\end{enumerate}
	
	Overall, we represent machine self-awareness as $MA=\{\mathbb{C},\mathbb{S},G_K,G_T,\mathbb{Q}\}$, which serves as a central controller and provides global guidance to the self-directed  learning process.
	
	\subsubsection{External Awareness}  External awareness represents machines' perception abilities towards the physical world and how the machine is seen by the external world. 
	We represent external awareness using the  following key components. 
	\begin{itemize}
		\item 
		\textit{Relations}. 
		The physical world consists of entities. 
		Different  from previous works which mostly represent entities  using texts, 
		we define the  set of entities  $\mathcal{V}=\{V_{1}, V_{2}, ...\}$ in a multi-modal way.
		Each entity is defined as $V_i = \{V_i^j\}^{M}_{j=0}$, where $M$ is the number of  modalities. The entities in $\mathcal{V}$ are interrelated. 
		We define their relations $\mathcal{R}$ in a recursive and hierarchical way. For example, the relation ``is grandfather of'' could be defined recursively by stacking the relation ``is father of'' twice.
		Formally, we have some basic relations in $\mathcal{R}$ first. We perform statistical inference on  basic relations to induce new relations and add statistically  significant ones to $\mathcal{R}$. 
		\item \textit{Knowledge}. 
		Humans rely on commonsense and domain knowledge to perform tasks. 
		We define    knowledge as $G_K = \{G_C, G_D\}$, where $G_C$ denotes commonsense knowledge and $G_D$ represents domain knowledge. Both $G_C$ and $G_D$ could be represented as graphs, where vertices represent concepts $V$ and edges represent relations $R$. While commonsense knowledge remains roughly the same in SDML, domain knowledge may be updated frequently during the learning process.
	\end{itemize}
	
	
	\subsubsection{Construction of self-awareness}
	
	We construct self-awareness by combining internal awareness and external awareness. To initialize the values of internal awareness, we collect information including computation speed and memory size of machines, task conditions on inputs, system constraints, et. To initialize the values of external awareness, we collect information including entities in the external world and their relations, seed commonsense knowledge such as ConceptNet, domain knowledge, etc. During the SDML process, we leverage  feedback collected from learning outcomes to update certain parts of self-awareness, including state, relationship between tasks, domain knowledge, etc. The update of self-awareness is conducted in a stable manner without catastrophic forgetting. Internal and external awareness are updated in a joint way instead of greedily to pursue globally optimal updates.

	
	
	
	\subsection{Self-directed Machine Learning Process}
	\label{sec:process} 
	With the guidance of our proposed self-awareness, machine algorithms can conduct self-directed learning processes. As shown in Fig.~\ref{fig:archi}, we propose an  implementation of a self-directed machine learning process. When interacting with the external dynamic environment, SDML can select learning tasks, data, models, optimization strategies and evaluation metrics 
	in a self-directed way based on the guidance of self-awareness where the results will in turn provide feedback to improve self-awareness, 
	thus distinguishing SDML from the conventional machine learning paradigm. 
	Next, we will describe each component in detail, i.e., \textit{self task selection}, \textit{self data selection}, \textit{self model selection}, \textit{self optimization strategy selection} and \textit{self evaluation metric selection}.
	
	\subsubsection{Self Task Selection}
	
	Humans could usually decompose a complex ultimate goal into several fine-grained tasks, which may be preconditioned or helpful to the final goal. For example, if we plan to cook a sumptuous meal, we will decompose this ultimate goal into the following tasks: buying the ingredients, preparing the ingredients, cooking and seasoning. Each task could be split to even smaller tasks, and we will pay attention to the task which we perform the worst to lift our ability towards the ultimate goal to a large extent.
	
	Inspired by this, SDML is designed to decompose a complex goal into a sequence of fine-grained subtasks automatically. 
	SDML can choose a subset of learning tasks and determine the order of executing them under the guidance of self-awareness. Formally, we define the ultimate goal as $T^* = \{T_C^*, T_O^*, T_E^*, T_S^*\}$. Note that the graph of machine learning tasks $G_T$ are included in machine self-awareness. Thus, given $T^*$, $G_T$, machine capacity $\mathbb{C}$, machine state $\mathbb{S}$ and evaluation results $O$, SDML is able to develop a function $f_T$ that generates a task sequence $\mathbb{T} = \{T_i\}_{i=0}^{N_T} = f_T(T^*, G_T, \mathbb{C}, \mathbb{S}, O)$, where the $N_T$ tasks are selected from $G_T$ and could be duplicated. This function may be modeled by similarity matching among the inputs, outputs, evaluation metrics, and experimental outcomes of different tasks. 
	The selected tasks affect the selection of data, model and optimization strategies.
	
	\subsubsection{Self Data Selection}
	Humans have the ability to find 
	the materials most related to a given task for learning, so that we could improve our ability on the given task in a rapid and effective way. 
	Humans usually choose suitable materials to learn for the given problem, and SDML should have the similar ability to select proper data to learn with awareness of the ultimate goal, the selected task and its current state.
	
	Formally, given the task sequence $\mathbb{T}$,  machine state $\mathbb{S}$, machine capacity $\mathbb{C}$, evaluation results $O$, and the ultimate goal $T^*$, SDML selects the most suitable dataset for each task. That is, $\mathbb{D} = \{ D_i \}_{i=1}^{N_T} = f_D(T^*, T_i,\mathbb{C}, \mathbb{S},O ) $. The data selection process not only relies on the selected task but also the ultimate goal for better efficiency since it aims to choose data most related to the final target while discarding unrelated, noisy or even harmful data. 
	Therefore, the SDML is designed to improve itself in an efficient and rapid way. The data selection results will also influence the model design and optimization strategy selection. 
	
	\subsubsection{Self Model Selection}
	
	Humans are able to locate the potential candidate solutions given different learning tasks.
	Similarly, after adjusting the learning tasks, SDML is required to choose the learning models for each task. Formally, SDML designs models $\mathbb{M} = \{M\}_{i=0}^{N_T} = f_M(T_i, D_i, \mathbb{C}, \mathbb{S}, O)$ based on the task sequence $\mathbb{T}$, machine capacity $\mathbb{C}$, machine state $\mathbb{S}$ and evaluation results $O$. The design of learning models also influences the selection of optimization strategies, while the selected models together with the corresponding model performance may give feedback to the task selection component.
	
	\subsubsection{Self Optimization Strategy Selection}
	
	As people vary in their abilities and available learning time, each person has her own learning pace and style. Therefore, people tend to choose strategies that are most suitable for themselves. Inspired by this, SDML chooses optimization strategies under the guidance of machine self-awareness. To date, many optimization strategies are proposed by researchers in order to reduce human intervention in data, supervision, losses, and optimization. 
	The set of optimization strategies are encoded and updated in self-awareness, which affects the learning speed, degree and cost etc. The optimization strategies are most flexible in SDML, which can be set according to the tasks and models. The set of strategies could also be a key factor for selecting tasks and designing models. Formally, the chosen optimization strategies are modeled through considering tasks, models as well as machine self-awareness: $\mathbb{P} = F_P(\mathbb{T}, \mathbb{M}, MA)$. Finally, the selected optimization strategies, outcomes and evaluations will provide feedback to further update all roles in the self-directed learning process.
	
	\subsubsection{Self Evaluation Metric Selection}
	It is essential to set proper evaluation metrics in machine learning. In conventional machine learning, one or more evaluation metrics are set through manual design. However, there exist various evaluation metrics and even researchers are not clear which metrics should be used under different circumstances. Besides, different evaluation metrics often lead to different optimization directions during model training. We expect SDML to make evaluation in a self-directed way, e.g., choosing evaluation metrics adaptively from a large set of candidates or even develop new evaluation metrics. When solving a complex goal without clearly well-defined objectives, machines will achieve lifelong self-improvement in the world upon successful self evaluations.
	
	\subsection{Mathematical Formulation of SDML}
	In this section, we present a mathematical  formulation of self-directed machine learning, by integrating the elements and processes introduced in earlier sections. 
	Specifically, we propose a multi-level optimization based framework to formulate  SDML. 
	In this framework, there are multiple joint optimization problems, each corresponding to one process outlined in Section \ref{sec:process}.  
	These processes are organized into a directed acyclic graph (DAG). If there is a directed edge from process $A$ to process $B$, then $B$ is dependent on $A$: specifically, the optimal solution of $A$'s optimization problem is used as a variable in $B$'s optimization problem. These optimization problems are organized into six levels. 
	
	At the first level, we construct self-awareness, by solving the following optimization problem:
	\begin{equation}
		B^*(M)=\textrm{argmin}_{B}\; L_{sac}(B,M).
	\end{equation}
	where $B$ denotes self-awareness, $L_{sac}$ denotes a self-awareness construction loss, and $M$ are meta parameters. The optimization is conducted over $B$. $M$ is tentatively fixed at this stage and will be updated later on. Note that the optimal solution $B^*$ is a function of $M$ since $B^*$ is a function of the loss function which is a function of $M$.

	At the second level,  we perform task selection. Given the constructed self-awareness $B^*(M)$,  a set of candidate tasks $\mathcal{T}=\{t_n\}_{n=1}^N$ and a textual description $E$ of the target application, the goal of task selection is to select a subset of tasks $\mathcal{S}\subseteq \mathcal{T}$ and form them into a DAG. The task DAG specifies the dependency between tasks and the execution order of tasks. 
	At this stage, we solve the following optimization problem: 
	\begin{equation}
		\mathcal{S}^*(B^*(M))=\textrm{argmin}_{\mathcal{S}\subseteq \mathcal{T}}\; L_{ts}(\mathcal{T},\mathcal{S},E,B^*(M)).
	\end{equation}
	$L_{ts}$ is a task selection loss specifying the criteria of how to select the optimal subset of tasks and form them into a DAG. It is defined on the entire set of tasks $\mathcal{T}$, a candidate subset of tasks $\mathcal{S}$, the application description $E$, and self-awareness $B^*(M)$. The optimization is conducted over  $\mathcal{S}$.  
	
	At the third level, there are two processes: one is training data selection; the other is model selection. Data selection is defined as follows. For each task $s(M)$ in the selected task  subset $\mathcal{S}^*(B^*(M))$, from the training data $D_{s(M)}$ of task $s(M)$, we select a subset of training examples $C_{s(M)}\subseteq D_{s(M)}$. Data selection for task $s(M)$ amounts to solving the following optimization problem:
	\begin{equation}
		C^*_{s(M)}=\textrm{argmin}_{C_{s(M)}\subseteq D_{s(M)}}\; L_{ds}(D_{s(M)},C_{s(M)}).
	\end{equation}
	$L_{ds}$ is a data selection loss specifying the criteria of how to select the optimal subset of training data. It is defined on the entire set of training data $D_{s(M)}$ of task $s(M)$ and a candidate subset $C_{s(M)}$ of data examples. The optimization is conducted over $C_{s(M)}$. 
	
	Model selection is defined as follows. For each task $s(M)$ in the selected task subset $\mathcal{S}^*(B^*(M))$, given the search space of  architectures and hyperparameters of the model used to perform the task  $s(M)$, we select the optimal architecture and hyperparameters. The corresponding optimization problem is:
	\begin{equation}
		A^*_{s(M)}=\textrm{argmin}_{A_{s(M)}}\; L_{ms}(A_{s(M)},s(M)).
	\end{equation}
	$L_{ms}$ is a model selection loss specifying the criteria of setting the optimal architecture and hyperparameters. $A_{s(M)}$ represents candidate architectures and hyperparameters in the search space. Optimization is conducted over $A_{s(M)}$.   
	
	At the fourth level, there is a single process, which is optimizer selection. For each selected task $s(M)$, given a set of candidate optimizers $\mathcal{O}=\{o_n\}_{n=1}^P$, we select the optimal one $o^*_{s(M)}\in \mathcal{O}$ to train the selected model on the selected data. The corresponding optimization problem is:
	\begin{equation}
		o^*_{s(M)}=\textrm{argmin}_{o_{s(M)}\in \mathcal{O}}\; L_{os}(o_{s(M)},A^*_{s(M)},C^*_{s(M)}).
	\end{equation}
	$L_{os}$ is a loss specifying the criteria of selecting the best optimizer. The selection of the optimizer depends on the selected model $A^*_{s(M)}$ and selected data $C^*_{s(M)}$. 
	
	At the fifth level, there is a single process, which is to train weight parameters of the selected  model on the selected data, using the selected optimizer, which amounts to solving the following optimization problem: 
	\begin{equation}
		W^*_{s(M)}=\textrm{argmin}_{W}\; L_{wt}(W_{s(M)},A^*_{s(M)},C^*_{s(M)},o^*_{s(M)}).
	\end{equation}
	
	At the sixth level, there is a single process, which is to evaluate models  trained at the fourth stage on a validation set $F$. Meta parameters $M$ are updated by minimizing the validation loss, which amounts to solving the following optimization problem:
	\begin{equation}
		\textrm{min}_{M}\;  L_{val}(\{W^*_{s(M)}|s(M)\in S^*(M)\},F),
	\end{equation}
	where $L_{val}$ is the validation loss. 
	
	Putting these pieces together, we have the following multi-level optimization problem. 
	\begin{equation}
		\begin{array}{l}
			\textrm{min}_{M}\;  L_{val}(\{W^*_{s(M)}|s(M)\in S^*(M)\},F),\\
			s.t. \; W^*_{s(M)}=\textrm{argmin}_{W}\; L_{wt}(W_{s(M)},A^*_{s(M)},C^*_{s(M)},o^*_{s(M)}),\\
			\quad \;  o^*_{s(M)}=\textrm{argmin}_{o_{s(M)}\in \mathcal{O}}\; L_{os}(o_{s(M)},A^*_{s(M)},C^*_{s(M)}),\\
			\quad \; A^*_{s(M)}=\textrm{argmin}_{A_{s(M)}}\; L_{ms}(A_{s(M)},s(M)),\\
			\quad \;  C^*_{s(M)}=\textrm{argmin}_{C_{s(M)}\subseteq D_{s(M)}}\; L_{ds}(D_{s(M)},C_{s(M)}),\\
			\quad \; \mathcal{S}^*(B^*(M))=\textrm{argmin}_{\mathcal{S}\subseteq \mathcal{T}}\; L_{ts}(\mathcal{T},\mathcal{S},E,B^*(M)),\\
			\quad \;
			B^*(M)=\textrm{argmin}_{B}\; L_{sac}(B,M).
		\end{array}
	\end{equation}


	\section{Case Studies and Potential Applications}
	\label{sec:case}
	In this section, we showcase how to leverage the proposed self-directed learning framework to solve practical problems, in two case studies: 1) autonomous rescue robotics, and 2) automated computer programming. Besides, we discuss a few other potential applications of SDML.
	\subsection{Case Study I: Autonomous Rescue Robotics}
	
	\subsubsection{Problem Definition}
	In robotics applications such as autonomous search and rescue, it is crucial for the agents to derive analogous solutions based on learned knowledge, such as opening a window based on the learned skills of opening a door. This requires the agents to perform analogical reasoning, including understanding which jobs (e.g., manipulation,  locomotion,  navigation,  assembly, etc.) are analogous, adapting the actions of performing source jobs to an analogous target job, etc. Existing research on analogical reasoning~\cite{gentner2012analogical,sunstein1993analogical} heavily requires humans to manually build knowledge bases about the analogy relationships between jobs and to manually craft symbolic systems for adapting actions between analogous jobs, which is labor-intensive, expensive, difficult to evolve over time, and less robust. In existing works on job planning~\cite{galindo2008robot,cambon2009hybrid}, given a novel job, a software program (e.g., policy, job plan, PDDL~\cite{aeronautiques1998pddl} description, etc.) operating on a robotic system needs to be written by human experts to execute this job, which is time-consuming and not scalable. There have been a few data-drive approaches~\cite{dantam2018incremental,grover2020radar} aiming to reduce the dependency on humans. However, they require a lot of experts-provided annotations for model training and such annotations are difficult to obtain practical robotics systems.
	\subsubsection{An SDML-based Solution}
\begin{figure}[!ht]
		\centering
		\includegraphics[width=1.0\linewidth]{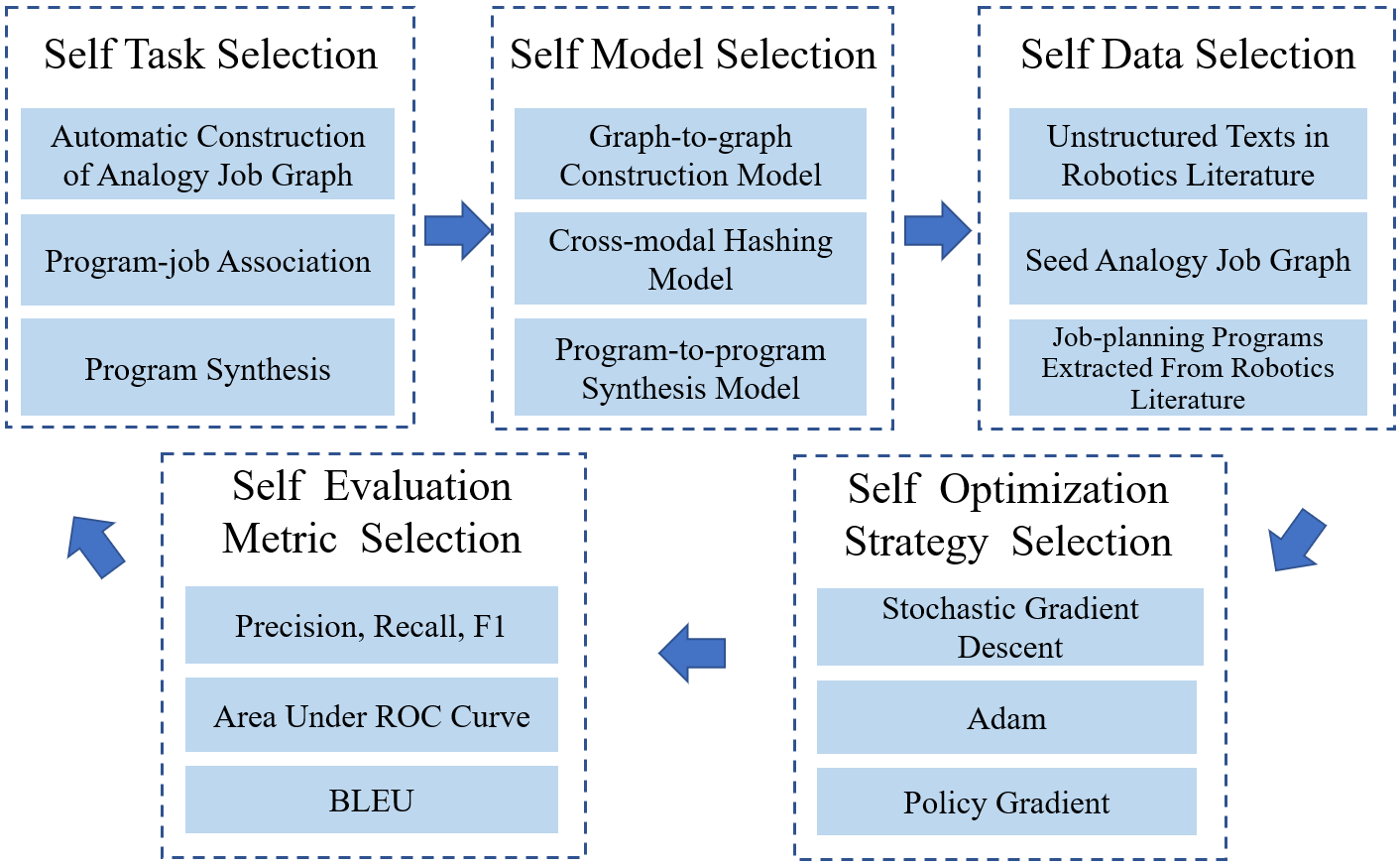}
		\caption{Illustration diagram of autonomous rescue robotics.}
		\label{fig:case1}
	\end{figure}

	To address the limitations of existing works, we can leverage our proposed SDML framework to develop  analogical reasoning systems (shown in Figure \ref{fig:case1}) that enable autonomous agents to master a much wider range of jobs without heavily relying on humans to provide supervision. Given a large set of analogous jobs, once the agent learns to solve one of them, our system enables it to automatically figure out how to solve the rest. This will make the autonomous agents more adaptive, autonomous, robust, and intelligent. Specifically, our solution aims to  automatically synthesize a correct and efficient program (e.g., an application domain definition written by the Planning Domain Definition Language~\cite{aeronautiques1998pddl}, which specifies the sequence of actions needed to accomplish a job and their preconditions and effects) to execute a previously unseen job by drawing analogies with previously seen jobs.

	\paragraph{Self task selection}
	\begin{itemize}
		\item Automatic construction of 
		analogy job graph: automatically build an analogy job graph (AJG) including jobs with analogy relationships and programs that can be used to perform these jobs. To perform analogical reasoning, we first need to know which jobs are analogous to each other. The AJG needs to be constructed automatically  without heavily relying on human annotations.
		\item Program-job association: for each job in the AJG, we aim to extract a program (represented using a syntax tree containing actions with preconditions and effects) from the robotics literature where the program can be used to perform this job.
		\item Program synthesis: given the AKB,  synthesize a target program to execute a novel job from the source programs of analogous jobs in the AKB. 
	\end{itemize}

	\paragraph{Self model selection}
	
	\begin{itemize}
		\item Graph-to-graph (G2G) construction model. The model is used for automatic construction of AJG. It takes unstructured texts in the robotics and job-planning literature and a small-sized seed AJG (created by domain experts) as inputs and constructs a more comprehensive graph containing previously unseen jobs and analogy relations. 
		The G2G model constructs a group of inter-correlated jobs and analogy relations collectively as a graph. In the construction process, high-order reasoning is performed to construct jobs having unobvious analogy relations. 
		To train the G2G model, we construct training examples, each containing a graph $G$ sampled from the seed graph and a subgraph $S$ of $G$, where the goal is to construct $G$ from $S$. We use the jobs and analogy relations in $S$ as queries to retrieve relevant paragraphs $P$ from the unstructured texts. Then we use a graph neural network to encode $S$ and use a BERT model trained on the entire collection of unstructured texts to encode $P$. Then the graph embedding of $S$ and BERT embedding of $P$ are fed into a two-layer hierarchical long short-term memory (LSTM)  network to construct new nodes. Given the constructed nodes, a Siamese network is used to construct edges: two jobs are connected if they are analogous. Given the constructed graph, we compare it with the ground-truth graph $G$ using graph edit distance. Since the objective is not end-to-end differentiable, we use policy gradient  to learn the weight parameters, where the reward is defined based on the graph edit distance.
		\item Cross-modal hashing model. We leverage a program extractor which analyzes the robotics literature and extracts all programs used for job planning. The number of extracted programs and the number of jobs in the AJG can be very large, which incurs huge computational costs when associating programs to jobs. We leverage a cross-modal hashing model to address this issue. We use a graph convolutional network to encode the AJG and learn a latent representation vector (whose elements are probability values) for each job in the graph. From the representation vector of each job, a hash code is sampled to represent this job. For programs extracted from robotics literature, we represent them using syntax trees and encode these trees using tree-structured LSTM networks. A hash code is sampled from the encoding of each program to represent this program. Given the hash codes of jobs and the hash codes of programs, we associate programs to jobs by calculating the Hamming distance between their hash codes. Hamming distance can be calculated extremely efficiently in memory, which enables the association to be done with minimal latency.
		\item  Program-to-program synthesis model. Given the learned analogy job graph where each job is associated with a program that can execute this job, we leverage it for analogical job planning. In previous job-planning approaches, a program (e.g., a PDDL  description) needs to be pre-defined by human experts before the robot can start its job. If a robot encounters a previously unseen job where a program has not been written for, the robot is unable to conduct this job. We leverage a program-to-program (P2P) synthesis  model  to solve this issue. The P2P model automatically generates a program for a novel job $J$ by leveraging programs pre-defined for jobs that are analogous to $J$. Specifically, from the analogy job graph, the P2P model retrieves the neighboring jobs that are analogous to $J$. The program associated with each retrieved job is represented using a syntax tree, which is encoded using a tree-structured LSTM  model. Then these program encodings are fed into a tree-structured LSTM decoder to generate the program for executing $J$.
	\end{itemize}
	
	For the graph-to-graph construction model, the following aspects are included in the model selection decision space: architecture of the graph neural network, architecture of the BERT model, architecture of the hierarchical long short-term memory network, and architecture of the Siamese network. 
	
	For the cross-modal hashing model, the following aspects are included in the model selection decision space: architecture of the graph convolutional network, architecture of the tree-structured LSTM network, and dimension of the hash codes. 
	
	For the program-to-program synthesis model, the following aspects are included in the model selection decision space: architecture of the tree-structured LSTM encoder, architecture of the tree-structure LSTM decoder, and number of hidden units in LSTM networks. 
	
	\paragraph{Self data selection}
	\begin{itemize}
		\item Data for training the graph-to-graph construction model: 1) unstructured texts in the robotics and job-planning literature; 2)  a small-sized seed AJG (created by domain experts) as inputs.  We construct training examples in the following way: sample a graph $G$ from the seed graph, sample a subgraph $S$ from $G$. Jobs and analogy relations in $S$ are used as queries to retrieve relevant paragraphs $P$ from the unstructured texts. The $(S,P,G)$ tuple forms a training example where $S$ and $P$ are the inputs and $G$ is the output.  
		\item Data for training the cross-modal hashing model: 1)  programs used for job planning, extracted from  
		robotics literature; 2) jobs in the AJG. We utilize a program extractor to analyze the robotics literature and extract all programs used for job planning. 
		\item Data for training the program-to-program synthesis model: jobs in the AJG and programs associated with these jobs. 
	\end{itemize}
	
	For each training example $x$, we automatically learn a weight $a\in[0,1]$. If $a$ is close to 1, it means that $x$ tends to be selected; otherwise, $x$ tends to be excluded. $a$ can be parameterized using a deep neural network which takes a latent representation of $x$ as input and produces a scalar between 0 and 1.
	
	\paragraph{Self optimization strategy selection} In the aforementioned tasks, some of them have differentiable objective functions and some do not. For differentiable objectives, we can use gradient based methods for optimization; and for non-differentiable ones, we can resort to reinforcement learning algorithms. 
	The candidate optimizers include: stochastic gradient descent, Adam, AdaGrad, RMSProp, policy gradient, deep deterministic policy gradient, actor-critic, asynchronous advantage actor-critic, trust region policy optimization, proximal policy optimization, and Q-learning. 
	
	\paragraph{Self evaluation metric  selection} Each of the aforementioned models can be evaluated from multiple perspectives.  The candidate evaluation metrics include precision, recall, F1, area under ROC curve, accuracy, BLEU, NIST, perplexity, etc.


	\subsection{Case Study II: Automated Computer Programming}
	\subsubsection{Problem Definition}
\begin{figure}[!ht]
		\centering
		\includegraphics[width=1.0\linewidth]{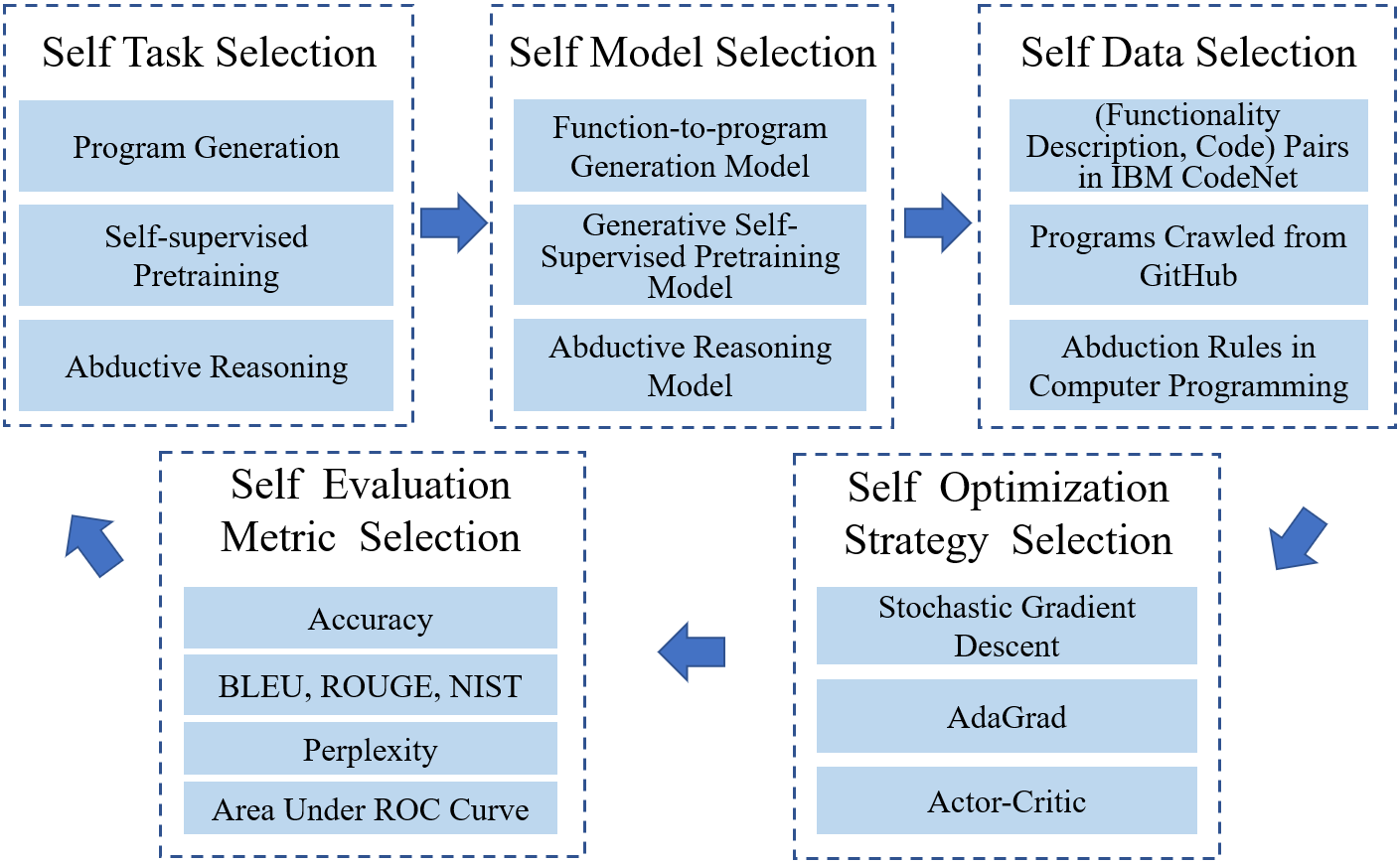}
		\caption{Illustration diagram of automated computer programming.}
		\label{fig:case2}
	\end{figure}

	Developing AI systems to automatically write computer-executable programs has attracted much research attention recently. Given a  text describing a target functionality to implement, an automated programming system takes the textual description as input and automatically generates a program to execute the function. Existing works for automated programming require a lot of training data which is difficult to obtain and do not perform reasoning to improve semantic correctness of generated programs. We aim to leverage our proposed self-directed ML framework to address these two problems, as shown in Figure~\ref{fig:case2}. 
	\subsubsection{An SDML-based Solution}
	To apply SDML for automated programming, we configure the following elements and processes. 
	
	\paragraph{Self task selection}
	\begin{itemize}
		\item Program generation: given a textual description describing a functionality, generate a preliminary program that can fulfill this functionality. 
		\item Self-supervised pretraining: pretrain the program generator to alleviate overfitting.
		\item Abductive reasoning: perform abductive reasoning to improve semantic correctness of generated programs. 
	\end{itemize}
	\paragraph{Self model selection}
	\begin{itemize}
		\item Function-to-program (F2P) generation model. It takes a functionality description $f$ as input and generates a program $p$ that can fulfill this functionality. The program is represented using a sequence of  constituent parse trees where on the nodes are symbols. The F2P model is based on an encoder-decoder architecture. The encoder takes the functionality description $f$ as input and generates a latent embedding for $f$. We use BERT as the encoder. The decoder takes the embedding of $f$ as input and generates a program. The architecture of the decoder is a sequence-of-trees long short-term memory (LSTM)  network. Given the embedding, the decoder first uses a sequential LSTM to decode a sequence of hidden states, each corresponding to one constituent tree. Then each hidden state is fed into a top-down tree-structured LSTM  to generate the corresponding constituent tree. 
		\item Generative self-supervised pretraining model. 
		To train the complicated F2P model, a lot of (functionality description, program) pairs are needed, which is difficult to obtain. Without sufficient training data, the F2P model is prone to overfitting. To address this problem, we leverage a  generative self-supervised pretraining approach to alleviate overfitting: we define a generative SSL task to learn a better decoder of programs. Given an original program  represented as a tree, we perform tree-alteration operations (e.g., removing a node, inserting a new node, moving a subtree of one node to another node, swap a parent node with its child node, etc.) to create a sequence of augmented programs $A_1,\cdots,A_K$.  Then we define the generative SSL task as: given $A_K$, 
		predict the reverse sequence of programs (including the augmented ones and the original one) $A_{K-1},A_{K-2},\cdots,O$ from which $A_K$ is generated. We develop an encoder-decoder model to perform this prediction task where an encoder is used to encode $A_K$ and a decoder is used to decode the sequence of programs $A_{K-1},A_{K-2},\cdots,A_1,O$. We use a bottom-up tree-structured LSTM network as the encoder which produces a latent embedding for $A_K$. Then the embedding is fed into a sequence-of-trees LSTM decoder~\cite{xie2017constituent} to generate a sequence of trees, each corresponding to a program in $A_{K-1},A_{K-2},\cdots,A_1,O$. \item Abductive reasoning model. Programs generated by the F2P model may contain semantic errors and fail to execute. To address this problem, we perform abductive reasoning on the generated programs to revise them. The abductive reasoning model  consists of a set $R$ of abduction rules. Each rule takes a set of statements as inputs and produces a new statement. Given the statements $S$ in a program generated by the F2P model, the abductive reasoning model selects a subset $T$ of statements from $S$, selects an abduction rule $r\in R$, and applies $r$ to $T$ to generate a new statement $s$, which is added to $S$. This procedure repeats several times until enough new statements are generated. Now the learning problem is which abduction rules should be applied to $T$. We develop an end-to-end reinforcement learning based approach  to perform this learning task. This RL-based framework learns an abductive reasoning policy (ARP) network which takes a set $S$ of statements and a set $R$ of abduction rules as inputs and produces a new statement. The ARP network is composed of three sub-networks: a statement selection (SS) network, an abduction rule selection (ARS) network, and a termination network. The SS network takes $S$ as input and selects a subset $T$ of statements. The ARS network takes $R$ and $T$ as inputs and selects a rule $r$ from $R$. Then the rule $r$ is applied to $T$ to infer a new statement. Note that it could be possible that $r$ is not compatible with $T$, meaning that there is no way to perform inference of $r$ on $T$. In this case, a negative reward will be assigned to guide the SS and ARS networks not to select invalid $T$ and $r$. The termination network takes the initial set of statements and the newly generated statements as inputs and produces a binary variable indicating whether the process of generating new statements should stop. The three sub-networks work together as follows. First, the termination network determines whether the generation process should continue. If so, the SS network selects $T\subseteq S$. Then the ARS network selects $r\in R$. Afterwards, $r$ is applied to $T$ to infer a new statement $s$, which is added to $S$. This procedure continues until the termination network determines to stop it. 
	\end{itemize}
	
	For the function-to-program generation model, the following aspects are included in the model selection decision space: architecture of the BERT encoder, dimension of  functionality  embeddings, architecture of the sequence-of-trees LSTM decoder, and Attention mechanism. 
	For the generative self-supervised pretraining model, the following aspects are included in the model selection decision space:  program augmentation policies, architecture of the bottom-up tree-structured LSTM encoder, architecture of the sequence-of-trees LSTM decoder, and dimension of program embeddings.  For the abductive reasoning model, the following aspects are included in the model selection decision space: 1) architecture of the statement selection network, architecture of the abduction rule selection network, and architecture of the termination network.

	\paragraph{Self data selection}
	\begin{itemize}
		\item Data for training the function-to-program generation model: (functionality description, code) pairs in the IBM CodeNet dataset.
		\item Data for generative self-supervised pretraining:  programs crawled from GitHub.  
		\item Data for training the abductive reasoning model: 1) abduction rules in computer  programming; 2) 
		(functionality description, code) pairs in the IBM CodeNet dataset.
	\end{itemize}
	
	The data selection mechanism is the same as that described in the previous section.
	
	\paragraph{Self optimization strategy selection}
	This application involves both differentiable and non-differentiable objective functions, which can be optimized using gradient based methods and reinforcement learning methods. The candidate optimizers are the same as those listed in the previous section. 
	
	\paragraph{Self evaluation metric  selection}
	We evaluate each model using multiple metrics from different perspectives. 
	The candidate evaluation metrics are the same as those listed in the previous section.

	\subsection{Potential Applications}
	Our SDML framework can be broadly applied to a variety of applications in NLP, CV, data mining and multimedia etc.,  beyond the two case studies described above. Here we present some examples. 
	\begin{itemize}
		\item 
		I) Commonsense-grounded controllable story writing. Controllable story writing, which automatically writes stories given control factors such as sentiment and storylines, finds broad applications. To write  meaningful and informative stories, it is necessary for ML models to incorporate external commonsense knowledge. For example, given a control factor ``On a sunny day, we go for exercise." which is a storyline, a story writing model  would not be able to write an interesting and informative story such as ``Yesterday was a sunny day. We wanted to do some exercise. Sunny weather is good for hiking and hiking is a popular exercise in California. So we went hiking." without knowing the commonsense knowledge that hiking is an exercise and is preferable on sunny days. Developing a commonsense-grounded story writer has several technical challenges: 1) how to automatically or semi-automatically collect  commonsense knowledge? 2)  how to efficiently retrieve relevant commonsense when writing stories? 3) how to train highly-performant commonsense-grounded story writing systems when the size of the story corpus covering commonsense knowledge is limited? We leverage our proposed SDML framework to automatically solve these problems via self-directed selection of tasks, data, models, and optimizers. \\
		II) Controllable video captioning aims to control the video caption generation process by auxiliary information guidance, e.g., the stylistic label~(romantic, humorous)~\cite{gan2017stylenet}, POS~(Part-of-Speech) tag sequence~\cite{deshpande2019fast}, or one exemplar sentence~\cite{yuan2020controllable}. The most challenging one is leveraging the exemplar sentence, which means generating a corresponding video caption sharing the same syntactic structure with one exemplar sentence. For example, when the groundtruth caption is ``A group of people are dancing'' with the example sentence ``Bunch of green bananas hanging in front of a banana tree''. The model may output ``Group of young people dancing in front of a live audience''. So the model needs to extract the syntactic structure of the exemplar sentence, incorporate it into the caption generation process reasonably and avoid being distracted from the extra noisy semantic information. The difficulties can be summarized as follows: 1) how to effectively extract syntactic information for further caption generation with limited exemplar sentences. 2) how to preserve video semantics in the generated captions despite the disturbances from exemplar sentences. Therefore, the capability of the SDML framework in the self-directed selection of tasks, data, models, and optimizers can be utilized for addressing the challenges mentioned above.
		
		\item 
		I) Semantics-aware chatbots. In open-world dialog systems, especially goal-oriented dialog systems, it is very important to understand the semantics of conversation histories and perform reasoning on the semantics, in order to give  informative, correct, and useful responses. For example, given a human utterance ``I want to drink some coffee. How far is the coffee shop?", without understanding the semantics of this utterance, the chatbot tends to give an uninformative and boring response such as ``Sounds cool". In contrast, if using semantic parsers  to parse the query ``How far is the coffee shop?" into a logical form which represents the semantics of this query and performing reasoning on the logical form together with an external knowledge base, the chatbot is able to give a useful and informative response such as ``It’s about half mile." To develop semantics-aware chatbots, several  technical challenges need to be addressed: 1) given limited annotated (utterance, logical forms) pairs, how to train highly accurate semantic parsers? 2) given limited  (logical forms, response) pairs, how to train a semantics-aware response generation model that is resilient to overfitting? 3) how to train reasoning systems  to infer deeper semantics? Leveraging the capability of SDML in performing self  selection of tasks, data, models, and optimizers, the aforementioned challenges can be automatically coped with.  \\
		II) Video dialogues. This is also known as the task of audio-visual scene-aware dialog (AVSD)~\cite{alamri2019audio,das2017visual}, which requires an agent to hold the conversation with humans in natural, conversational language about video content and audio content in the format of dialogue box. In real world dialog systems, applied into daily use including tools for early childhood education, it is important to understand the semantics of conversation histories and the multi-modal representation of video, audio, and text, so as to give responses with high relevance as well as enough information. For example, given a video with audio about a person walking by a bag and leaving a book, and then given a question in natural human language "Does she walk quickly or slowly?, without understanding the video, the response could be random "Quickly" or "Slowly". While an agent who fully understands the video context can give the response "She walks pretty slowly back and forth before putting down the book", which obviously gives more information in a softer description~\cite{alamri2019audio,liu2020violin}. To develop audio-visual scene-aware dialog agents, several technical challenges need to be addressed: 1) how to train highly accurate semantic parsers? Since all the questions and responses are given in the textual format, precisely understanding the question is the key part. 2) how to train systems to fuse and understand multi-modal information including the dynamic scene, the audio, and the history (previous rounds) of the dialog? Leveraging the capability of SDML in performing self selection of tasks, data, models, and optimizers, the challenges mentioned above could be easily coped with.
	\end{itemize}

	\section{Related Works}
	\label{sec:related}
	In a section, we discuss several existing works that are related to our proposed SDML paradigms. In concrete, \textit{Curriculum Learning, Meta Learning, Automated Machine Learning}, \textit{Lifelong/Continual Learning} and \textit{Reinforcement Learning} are all related to SDML.
	
	
	
	

	\subsection{Curriculum Learning}
	
	Curriculum learning (CL)~\cite{bengio2009curriculum,wang2021survey} is a training strategy of machine learning that selects the most suitable examples or tasks (with adjustable loss weights) for each current training step, aiming to improve the model's generalization ability, robustness, convergence speed, etc. CL can be seen as a self-directed learning strategy, since the learning algorithm itself makes efforts to handle the biases (e.g., class imbalance, label noise, etc.) of the training set by autonomously re-weighting the training samples, which is also one of the most crucial advantage of CL~\cite{ren2018learning}. 
	
	Existing CL strategies can be roughly categorized into three groups: original CL, hard example mining (HEM), and automatic CL. 
	\textit{Original CL}~\cite{bengio2009curriculum} proposes to train the machine learning model with easier data subsets (or easier subtasks), and then gradually increase the difficulty level of data (or subtasks) until the whole training dataset (or the target task(s)). 
	Imitating human learning from easy to hard, CL helps to improve the performance on test set (or target tasks) and the convergence speed. 
	Predefined CL methods measure data difficulty with task-specific domain knowledge. For example, longer sentences are often supposed as harder training data in NLP tasks, and audios with lower Signal to Noise Ratio are expected to be more noisy and thus harder in speech recognition tasks. 
	On the other hand, self-paced learning (SPL), a primary branch of CL, takes the example-wise training loss of the current model as the criteria for difficulty measurement. SPL alternatively optimizes the re-weighting variables for each data example and the model parameters. Other methods also decide difficulty measurement by the losses of pretrained teacher models.
	\textit{Hard example mining (HEM)}~\cite{shrivastava2016training} is another well-studied and popular data selection strategy that proposes to assign higher weights to harder data at each training step, taking an opposite paradigm to CL. 
	The basic assumption of HEM is that the harder examples are more informative than easier examples and thus more beneficial for model learning. 
	The difficulty in HEM is often defined according to the current model losses on examples or the gradient magnitude. 
	\textit{Automatic CL} methods discard the prior assumptions on the data difficulty and training strategies and aim to learn the loss weights of data examples at each training step according to a specific target (e.g., higher training efficiency, higher validation/testing performance, etc.). 
	A typical way to achieve automatic CL is to make a reinforcement learning (RL) agent learn to assign weights to data. Concretely, the \textit{state} in RL is the data, current state of model, training epoch, etc., the \textit{action} in RL is to assign weights to the data, and the \textit{reward} is defined according to the requirement of tasks. Other automatic CL methods optimize the re-weighting strategies by Bayesian Optimization~\cite{tsvetkov2016learning}, meta-learning~\cite{ren2018learning}, gradient descent~\cite{jiang2018mentornet,kim2018screenernet}, adversarial learning~\cite{Zhang2020FewCostSO}, etc. 
	
	CL has been theoretically and practically proven an effective strategy to improve the model's robustness~\cite{chen2021curriculum2}, performance on target data/task, convergence speed~\cite{chen2021curriculum1}, etc. 
	It will be interesting to exploit more self-directed methodologies by letting the algorithm itself decide the most suitable loss functions (or learning objective), training data (with automated data generation), and hypothesis space for the model optimization.

	\subsection{Meta Learning}
	
	Last decade has witnessed a prosperous development for supervised learning, 
	which usually depends on large labeled datasets and trains a huge model with a large number of parameters from scratch. Thus, the requirement for data and computing resources is relatively high. However, there are many applications where data is difficult or expensive to collect, or computing resources are limited. Since the lack of training data, supervised learning is not suitable for these tasks and shows bad performances. 
	
	
	For the sake of human-like learning, meta-learning which targets at simulating the concept of ``learning to learn'', provides a paradigm where machine learning models are built based on experience with related tasks. Meta-learning has been becoming a very hot research topic in both academy
	and industry since the year of 2017, covering many research communities including machine learning, computer vision, natural language processing, data mining and multimedia. 
	
	We summarize meta-learning as a series of techniques that can learn prior experience across tasks in a systematic, data-driven manner. 
	We define the problem of meta-learning in two views, i.e., {\it task distribution view} and {\it learner and meta-learner view}.
	
	\textit{Task Distribution View}
	A good meta-learning method should help the model $f_\theta$ gain the ability of learning to learn across tasks and improve its performance on a distribution of tasks $p(\tau)$, including potentially unseen tasks. The optimal model parameters are:
	\begin{equation}\label{eq:meta_1}
		\theta^* = \mathop{\arg\min}_{\theta} \mathbb{E}_{\tau_i \sim p(\tau)}[\mathcal{L}_\theta(\tau_i)].
	\end{equation}
	
	To implement the optimization goal Eq.~\eqref{eq:meta_1}, we usually sample $M$ meta-train tasks from $p(\tau)$, with which we learn meta-knowledge $\omega$. The meta-knowledge $\omega$ guides the optimization of the model across tasks, and can have different meanings, such as parameter initialization, gradients, and optimization strategy. The training and validation sets of a meta-train task are often called support set and query set. Formally, we denote the set used in the meta-train stage as $\mathcal{D}_{meta-train}=\left\{(\mathcal{D}_{meta-train}^{\mathcal{S}}, \,\mathcal{D}_{meta-train}^{\mathcal{Q}}) ^{(i)}\right\}_{i=1}^M$. After meta-train stage, the model gets the optimal meta-knowledge $\omega^*$:
	\begin{equation}\label{eq:meta_2}
		\omega^* = \mathop{\arg\max}_{\omega} \mathrm{log}\:p(\omega|\mathcal{D}_{meta-train}).
	\end{equation}
	Similarly, we denote the $N$ sampled tasks used in the meta-test stage as $\mathcal{D}_{meta-test}=\left\{(\mathcal{D}_{meta-test}^{\mathcal{S}}, \,\mathcal{D}_{meta-test}^{\mathcal{Q}}) ^{(i)}\right\}_{i=1}^N$. We use the learned meta-knowledge to train the model on each previously unseen task and get the optimal model parameters:
	\begin{equation}\label{eq:meta_3}
		\theta^* = \mathop{\arg\max}_{\theta} \mathrm{log}\:p(\theta|\omega^*,\mathcal{D}_{meta-test}^{\mathcal{S}}).
	\end{equation}
	We can evaluate the meta-learning algorithm by the performance of $\theta^*(i)$ on the corresponding meta-test query set $\mathcal{D}_{meta-test}^{\mathcal{Q}}$.

	\textit{Learner and Meta-learner View} 
	Another common view is that meta-learning decomposes the process of parameter update into two stages: base-learning stage and meta-learning stage. As aforementioned, the dataset for each task $\tau_i \sim p(\tau)$is divided into support set $\mathcal{S}^{(i)}$ and query set $\mathcal{Q}^{(i)}$. During the base learning stage, an inner learner model $f_\theta$ is trained on the support set $\mathcal{S}^{(i)}$ for solving a given specific task. During meta-learning stage, an outer meta-learner $g_\omega$ is applied to improve  an outer objective $\mathcal{L}^{meta}$ that is calculated on the query set $\mathcal{Q}^{(i)}$. The outer optimization problem contains the inner optimization as a constraint. Using this notation, the parameter $\omega$ of meta-learner $g_\omega$ can be regarded as meta-knowledge. The meta-learning algorithm can be formulated as follows:  
	\begin{align}
		\omega^* &= \mathop{\arg\min}_{\omega} \sum_{i=1}^{N} \mathcal{L}^{meta}\left(\theta^{*(i)}(\omega), \omega, \mathcal{Q}^{(i)}\right) ,\label{eq:meta_4}\\
		s.t. \quad \theta^{*(i)}(\omega) &= \mathop{\arg\min}_{\theta} \mathcal{L}^{base}\left(\theta, \omega, \mathcal{S}^{(i)}\right) ,
		\label{eq:meta_5}
	\end{align}
	where $\mathcal{L}_{meta}$ and $\mathcal{L}_{base}$ refer to the outer and inner objective losses respectively, such as cross entropy in the case of few-shot classification.

	\subsection{Automated Machine Learning}
	Most machine learning methods have a plethora of design choices that need to be made beforehand, and their performance is shown to be very sensitive to these choices. Furthermore, the desirable choices of algorithm design often vary over different tasks and hence the algorithm configuration requires intensive expertise, which becomes a substantial hurdle for new users and further restricts the applicability and feasibility of modern machine learning methods in a wider range of public fields. 
	To remedy this issue, automated machine learning (AutoML) is developed to configure machine learning methods in a data-driven, object-oriented and automatic way. AutoML aims to learn the configuration of machine learning methods that attains the best performance on the specific task. In this way, AutoML largely reduces the background knowledge needed to customize modern machine learning methods in specific application domains, which makes machine learning technologies more user-friendly~\cite{he2020automl,yao2018taking}. 
	Complete AutoML pipelines have the potential to automate every step of machine learning, including auto data collection and cleaning, auto feature engineering, and auto model selection and optimization, etc. Due to the popularity of deep learning models, hyper-parameter optimization (HPO)~\cite{bergstra2011algorithms,wang2021explainable,liu2021meta} and neural architecture search (NAS)~\cite{guan2021autoattend,wei2021autoias,qin2021graph} are most widely studied. AutoML has achieved or surpassed human-level performance~\cite{zoph2017neural,liu2018darts,pham2018efficient} with little human guidance in areas such as computer vision~\cite{zoph2018learning,real2019regularized}.

	\subsection{Lifelong Learning/Continual learning}
	Lifelong learning models are designed by humans to obtain an ability to continually learn new skills and knowledge, while not forgetting what has been learned. This ability is already available to humans, but remains a challenge for computational systems and autonomous agents~\cite{lifelongReview}. In the lifelong learning approaches, the idea of
	self-directed learning is reflected in many places, such as selection of key parameters, pruning network architecture, and replaying memories.
	
	The existing methodologies for Lifelong learning can be divided into three categories: regularization-based methods, parameter isolation-based methods,  and replay-based methods. Regularization-based methods~\cite{ewc,gem} limit how far the parameters can move from values that were optimal for previous learning. They automatically determine which parameters are essential and then penalize the change to these parameters in future training. Parameter isolation-based methods~\cite{packetnet,learntogrow} autonomously dedicate different subsets of the model parameters to different skills and knowledge. As the model learns new things, the capacity of the model may increase as needed, while some redundant parameters may be reinitialized or pruned. Replay-based methods~\cite{icarl,ucir} need a memory component to store data from the previous learning process. The stored data are strategically sampled and help the model avoid forgetting in future learning.
	
	However, in lifelong learning/continual learning, what learning sequence to follow and what learning regime to use between different knowledge and skills are still not self-directed. In the future, making lifelong learning/continual learning work in a more and even complete self-directed way will be an interesting and exciting area.
	
	
	
	
	\subsection{Reinforcement Learning}
	
	
	
	Reinforcement Learning (RL), which emits a sequence of actions in the Markov Decision Process (MDP), is sometimes seen as a new category of machine learning method~\cite{haydari2020deep}. Other than having an explicit ground-truth target, RL models are usually optimized to maximize a long-term reward. In another word, RL tasks usually do not have an immediate target like ``this image should be an apple'' but have a long-term goal like ``win the game of Go with a sequence of actions''~\cite{silver2017mastering}. As both RL and our SDML are optimized towards a long-term goal and do not have explicit supervision, the two paradigms have a lot of features in common. But, there are also some subtle and crucial differences in between. In the following, we briefly introduce Reinforcement Learning considering these differences.
	
	Firstly, the optimization target. An RL model is usually designed for a specific task, e.g., chess~\cite{silver2017mastering}, games~\cite{mnih2013playing} robotics~\cite{kober2013reinforcement}, auto-driving~\cite{sallab2017deep}. Given a task, the current RL method would construct a model and sample data traces from the task data distribution. Recently, equipped with Deep Learning as a power function fitting tool, RL methods tend to leave most of the work to its underlying deep learning model and focus on the optimization methods. As a comparison the target of SDML is much bigger, which the target is far more than a specific task, also includes the process of finding a proper model (self data selection), data (self-data selection), and optimization method (self optimization strategy selection) to optimize the task. It would be possible to model the target of SDML as an RL target and optimize that target in the abstract. Still, it would not be technically possible because the target of SDML would be too complicated and comprehensive to be modeled with a single RL model.
	
	An RL model is usually optimized via the samples drawn by sampling. Early sampling strategies are inspired by the multi-arm bandit that tries to balance {\it exploration} and {\it exploitation}. 
	The observation in RL is more likely the ``data point'' in supervised learning, all the observations come from the same distribution, and the goal is to select the best ``action'' under that circumstance. While in SDML, we work with ``task distribution'' rather than ``data distribution''. The goal is more general and difficult, as there may be dozens of different data distributions that SDML is required to handle, and our observation is a task distribution rather than a single data point.

	From the perspective of modeling, RL agents could be categorized into model-free and model-based~\cite{thrun2000reinforcement}. The former is known for policy-gradient methods that directly learn the mapping from observation $\mathcal{O}$ to the best action $a$, which has achieved great success in early times~\cite{sutton2000policy}. The latter has been studied recently in~\cite{schrittwieser2020mastering, hafner2019dream}, where besides the policy is modeled, the environment is also modeled to predict the possible reward $R[a]$ given an action $a$. The modeling of reward helps model-based RL methods have dominated several RL areas. Modeling the environment as a ``reward function'' is an ingenious abstraction as the whole environment would be hard to model. In SDML, we have to model more than the ``reward function''. Usually, we humans have an understanding of the world so as to abstract commonsense information from different environments and make decisions for each of them. While in the scope of SDML, the environment could cover more: task distribution, model distribution, and even knowledge distribution. A SDML model has to model all these distributions at the same time so as to master them all.
	
	
	
	\section{Conclusions and Future Directions}
	\label{sec:concl}
	Drawing inspiration from  humans' self-directed learning, we introduce the principal concept and propose the framework of self-directed machine learning (SDML). SDML aims for high autonomy, which conducts self task selection, self data selection, self model selection, self optimization strategy selection as well as self evaluation metric selection instead of requiring humans to manually perform the selections. Our proposed SDML framework consists of key elements including internal awareness and external awareness and key processes including self-directed task selection, data selection, model selection, and optimizer selection. We propose a multi-level optimization based framework to formulate SDML. In the two case studies including autonomous rescue robotics and automated computer programming, we illustrate how to leverage SDML to solve complicated practical problems autonomously. 
	
	It is worth  noting that our proposed SDML framework is agnostic to specific ML applications and can be broadly applied to improve a variety of ML tasks including but not limited to: classification, regression,  clustering, text generation, dialog systems, machine translation, document summarization, object detection, semantic segmentation, visual question answering, time series prediction, link and node prediction in graphs, etc.  
	
	For future work, we plan to investigate the following research directions.
	\begin{itemize}
		\item \textbf{Interpretable SDML} Being reliable is almost a must for ML models to be willingly used by humans. To gain trust from humans, we will develop interpretable SDML methods which generate explainable and transparent predictions.  Most of the prior approaches for interpretable ML focus on finding out key evidence from the input data (such as phrases in texts and regions in images) that is most relevant to a prediction, then using this evidence to justify the meaningfulness of the prediction. 
		However, in many cases, the attributed evidence does not make sense to the human. A fundamental reason is that the reasoning processes of ML models and humans are not aligned, though they reach the same prediction outcome. To address this issue, we  plan to study weakly supervised model-interpretation. Specifically, we will develop natural language processing methods to analyze texts and automatically extract the decision-making processes of humans therefrom, then inject these structured processes into the SDML framework as an inductive bias to achieve human-machine alignment. As a longer-term goal, we  will collaborate with cognitive scientists to deeply understand the fundamental mechanisms of how humans interpret  phenomena and decisions and use these mechanisms to guide the design of SDML frameworks.  
		\item \textbf{Robust SDML} 
		In many applications such as healthcare, finance, etc.,   decisions are mission-critical. ML-aided  decision support software is required to be secure and robust against malicious attacks. The existing clinical ML models are shown to be vulnerable to adversarial examples. For example, given a chest X-ray that is predicted by a convolutional neural network as containing pneumonia, adding tiny perturbations (that are not perceivable by the human) could render the model thinks this image has no pneumonia. Most of the prior defense methods are highly customized to specific attacks; thus they may easily become futile when the attacks change. We will develop SDML frameworks which represent attacks and defenses in a unified way and accordingly devise defense techniques that are able to cope with various forms of attacks and robust to the changes of attacks.  As a longer-term goal, we will collaborate with cryptographers to develop ML-specific homomorphic encryption (HE) methods that allow SDML  training and inference on ciphertexts. 
		\item \textbf{Sample-efficient  SDML} In many applications, due to privacy concerns and  administrative regulations,  the amount of  data that the SDML framework could access for model training is usually quite limited. And the cost (regarding time and financial budget) of attempting to obtain more data  grows super-linearly with the amount of data. We are  interested in answering a research question: under circumstances where we do not have a large amount of data due to cost-control purposes, can we still be able to learn highly performant  ML models via SDML? This question has been investigated in previous studies based on few-shot learning, meta-learning, transfer learning, etc. However, these approaches do not perform reasoning to mitigate data deficiency. We plan to  bridge this gap. 
		We will develop neural-symbolic reasoning systems to understand the  relationship between high-level variables in  data, develop logical and inductive reasoning systems  to discover the causality between tasks, and leverage the reasoning outcomes to guide self-directed learning processes. We will develop graph neural network based reasoning systems to automatically discover logical rules and conduct long-range multi-step complex reasoning to determine the execution order and interactions between models.
	\end{itemize}

	\bibliographystyle{plain}
	\bibliography{sdml}
	
	\begin{IEEEbiography}[{\includegraphics[width=1in,height=1.25in,clip,keepaspectratio]{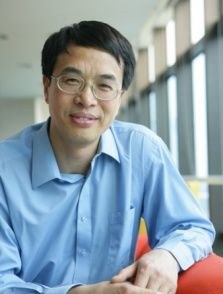}}]{Wenwu Zhu}
		is currently a Professor in the Department of Computer Science and Technology at Tsinghua University, the Vice Dean of National Research Center for Information Science and Technology. Prior to his current post, he was a Senior Researcher and Research Manager at Microsoft Research Asia. He was the Chief Scientist and Director at Intel Research China from 2004 to 2008. He worked at Bell Labs New Jersey as Member of Technical Staff during 1996-1999. He received his Ph.D. degree from New York University in 1996.
		
		His current research interests are in the area of data-driven multimedia networking and multimedia intelligence. He has published over 350 referred papers, and is inventor or co-inventor of over 50 patents. He received eight Best Paper Awards, including ACM Multimedia 2012 and IEEE Transactions on Circuits and Systems for Video Technology in 2001 and 2019.  
		
		He served as EiC for IEEE Transactions on Multimedia (2017-2019). He serves as the chair of the steering committee for IEEE Transactions on Multimedia, and he serves Associate EiC for IEEE Transactions for Circuits and Systems for Video technology. He serves as General Co-Chair for ACM Multimedia 2018 and ACM CIKM 2019, respectively. He is an AAAS Fellow, IEEE Fellow, SPIE Fellow, and a member of The Academy of Europe (Academia Europaea).
	\end{IEEEbiography}
	
	\begin{IEEEbiography}[{\includegraphics[width=1in,height=1.25in,clip,keepaspectratio]{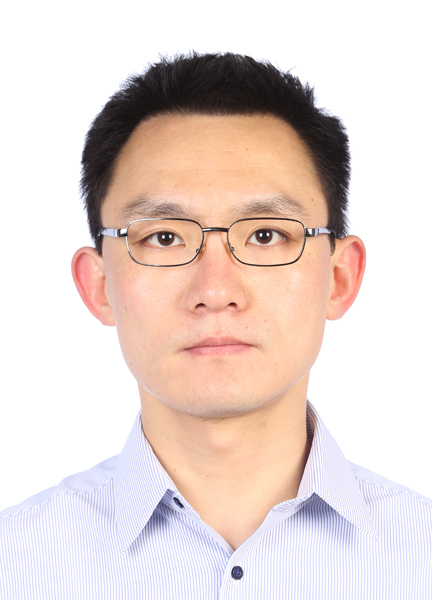}}]{Xin Wang}
		is currently an Assistant Professor at the Department of Computer Science and Technology, Tsinghua University. He got both of his Ph.D. and B.E degrees in Computer Science and Technology 
		from Zhejiang University, China. He also holds a Ph.D. degree in Computing Science from Simon Fraser University, Canada. 
		His research interests include cross-modal multimedia intelligence and inferable recommendation in social media. 
		He has published several high-quality research papers in top journals and conferences including TPAMI, TKDE, ICML, MM, KDD, WWW, SIGIR etc.
		He is the recipient of 2017 China Postdoctoral innovative talents supporting program. He receives the ACM China Rising Star Award in 2020.
	\end{IEEEbiography}
	
	\begin{IEEEbiography}[{\includegraphics[width=1.0in,height=1.25in,clip,keepaspectratio]{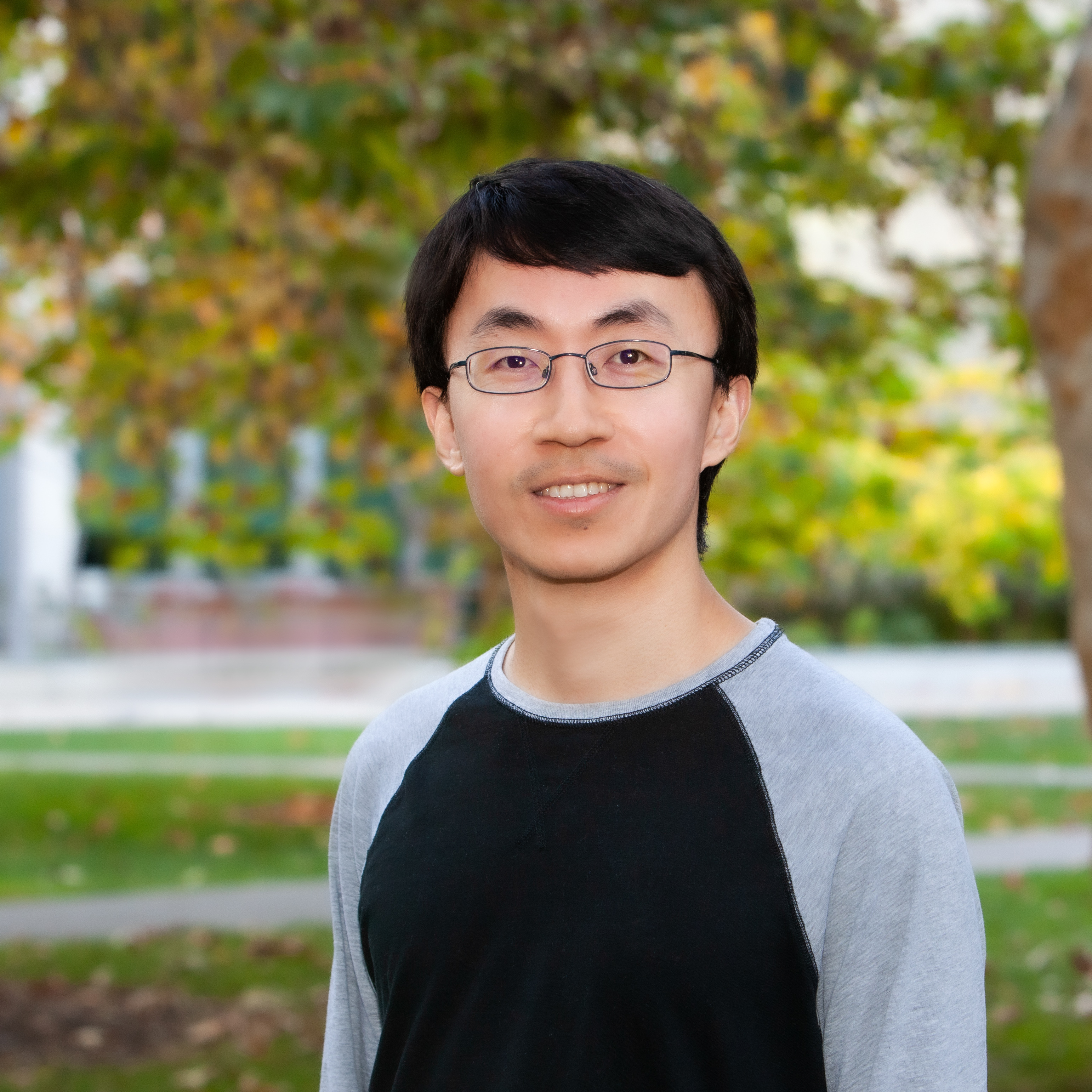}}]{Pengtao Xie}
		is currently an assistant professor of Electrical and Computer Engineering at University of California, San Diego. He obtained his PhD from the School of Computer Science at Carnegie Mellon University.  His research interests mainly lie in machine learning inspired by humans’ learning skills (especially classroom learning skills), with applications to neural architecture search, multi-level optimization, etc.
	\end{IEEEbiography}

\end{document}